\documentclass[letterpaper, 10 pt, journal, twoside]{IEEEtran}

\IEEEoverridecommandlockouts
\makeatletter

\let\proof\@undefined
\let\endproof\@undefined
\makeatother

\usepackage{amsmath}  
\usepackage{amssymb}  
\usepackage{amsthm}
\usepackage{amsfonts}
\usepackage{mathtools}
\usepackage{paralist}
\usepackage{verbatim}
\usepackage{color}
\usepackage{graphicx}
\usepackage{caption}
\usepackage{times}

\usepackage{algorithm} 
\usepackage{algpseudocode}

\usepackage[us]{datetime}
\usepackage{caption}
\usepackage[usenames,dvipsnames,table]{xcolor}

\usepackage{soul}

\usepackage{todonotes}
\presetkeys{todonotes}{inline}{}

\usepackage{subcaption}

\usepackage{latexsym}
\usepackage{color}

\usepackage[noadjust]{cite}

\usepackage{caption}
\usepackage{float}
\usepackage{textcomp} 

\usepackage{tikz}
\usetikzlibrary{shapes,fit,arrows,calc,intersections,
positioning,backgrounds,decorations.markings, patterns,external}

\tikzset{external/system call={latex \tikzexternalcheckshellescape -halt-on-error
    -interaction=batchmode -jobname "\image" "\texsource";
    dvips -o "\image".eps "\image".dvi;
ps2eps "\image.eps"}}

\let\labelindent\relax 
\usepackage[inline]{enumitem} 
\setenumerate[0]{label=\roman*.}

\usepackage{subfiles}
\usepackage{bm}
\usepackage[utf8]{inputenc}
\DeclareUnicodeCharacter{2212}{−}

\tikzset{
  connect/.style args={(#1) to (#2) over (#3) by #4}{
    insert path={
      let \p1=($(#1)-(#3)$), \n1={veclen(\x1,\y1)},
      \n2={atan2(\x1,\y1)}, \n3={abs(#4)}, \n4={#4>0 ?180:-180}  in
      (#1) -- ($(#1)!\n1-\n3!(#3)$)
      arc (\n2:\n2+\n4:\n3) -- (#2)
    }
  },
}

\usepackage{pgfplots}
\usepgfplotslibrary{groupplots,dateplot}
\pgfplotsset{compat=newest}
\pgfplotsset{every axis/.append style={font=\footnotesize}}

\usetikzlibrary{patterns, shapes.arrows, 3d, calc, angles, intersections, arrows.meta, positioning}
\usetikzlibrary{trees}
\usetikzlibrary{decorations.pathmorphing}
\usetikzlibrary{decorations.markings}
\usetikzlibrary{decorations.pathreplacing,decorations.pathmorphing}

\tikzset{%
  >={Latex[width=2mm,length=2mm]},
  base/.style = {rectangle, draw=black,
  minimum width=2.5cm, minimum height=0.7cm,
  text centered, font=\sffamily},
  start/.style = {base, line width=0.5mm},
  end/.style = {base, line width=0.5mm},
  process/.style = {base, line width=0.5mm},
  decision/.style = {base},
  noborders/.style = {},
  split/.style 2 args={pattern=north west lines, pattern color=cyan!90!blue, draw=black,
  minimum width=2.5cm, minimum height=0.7cm,
  text centered, font=\sffamily, line width=0.5mm, path picture={\fill[orange!38, sharp corners]([xshift=#1]path picture bounding box.south) -- ([xshift=#1]path picture bounding box.north) -- (path picture bounding box.north west) -- (path picture bounding box.south west) -- cycle;}},
  }

\makeatletter
\let\NAT@parse\undefined
\makeatother

\usepackage{hyperref} 

\setlist[itemize]{noitemsep, nosep}

\usepackage{siunitx}
\usepackage{ifthen}

\usepackage{booktabs}

\renewcommand{\baselinestretch}{0.98}
\usepackage{tablefootnote}
\usepackage[hang,flushmargin]{footmisc}

\makeatletter
\def  \input@path{{./../fig/},{./fig/}}
\makeatother

\usepackage{scalerel}
\usetikzlibrary{svg.path}
\definecolor{orcidlogocol}{HTML}{A6CE39}
\tikzset{
orcidlogo/.pic={
  \fill[orcidlogocol] svg{M256,128c0,70.7-57.3,128-128,128C57.3,256,0,198.7,0,128C0,57.3,57.3,0,128,0C198.7,0,256,57.3,256,128z};
  \fill[white] svg{M86.3,186.2H70.9V79.1h15.4v48.4V186.2z}
               svg{M108.9,79.1h41.6c39.6,0,57,28.3,57,53.6c0,27.5-21.5,53.6-56.8,53.6h-41.8V79.1z M124.3,172.4h24.5c34.9,0,42.9-26.5,42.9-39.7c0-21.5-13.7-39.7-43.7-39.7h-23.7V172.4z}
               svg{M88.7,56.8c0,5.5-4.5,10.1-10.1,10.1c-5.6,0-10.1-4.6-10.1-10.1c0-5.6,4.5-10.1,10.1-10.1C84.2,46.7,88.7,51.3,88.7,56.8z};
}
}
\newcommand\orcidicon[1]{\href{https://orcid.org/#1}{\mbox{\scalerel*{
\begin{tikzpicture}[yscale=-1,transform shape]
\pic{orcidlogo};
\end{tikzpicture}
}{|}}}}

\title{
  Mobile Manipulator for Autonomous Localization, Grasping and Precise Placement of Construction Material in a Semi-structured Environment
}

\author{
  Petr \v{S}tibinger$^{1\orcidicon{0000-0002-7662-9230}}$,
  George Broughton$^{2\orcidicon{0000-0003-0071-5834}}$,
  Filip Majer$^{2\orcidicon{0000-0002-4921-3360}}$,
  Zden\v{e}k Rozsyp\'{a}lek$^{1}$,
  Anthony Wang$^{3}$,
  Kshitij Jindal$^{3}$,
  Alex Zhou$^{4}$,
  Dinesh Thakur$^{4\orcidicon{0000-0001-5046-8160}}$,
  Giuseppe Loianno$^{3\orcidicon{0000-0002-3263-5401}}$,
  Tom\'{a}\v{s} Krajn\'{i}k$^{2\orcidicon{0000-0002-4408-7916}}$,
  Martin Saska$^{1\orcidicon{0000-0001-7106-3816}}$
  \thanks{$^{1}$The authors are with the Czech Technical University in Prague, Faculty of Electrical Engineering, Department of Cybernetics, Karlovo N\'{a}m\v{e}st\'{i} 13, 121 35 Prague, Czech Republic. {\tt\footnotesize email: \{stibipet, rozsyzde, martin.saska\}@fel.cvut.cz}.}
  \thanks{$^{2}$The authors are with the Czech Technical University in Prague, Faculty of Electrical Engineering, Department of Computer Science, Karlovo N\'{a}m\v{e}st\'{i} 13, 121 35 Prague, Czech Republic. {\tt\footnotesize email: \{brouggeo, majerfil, krajnt1\}@fel.cvut.cz}.}
  \thanks{$^{3}$The authors are with the New York University, Tandon School of Engineering, 5 MetroTech Center, 11201 Brooklyn NY, USA. {\tt\footnotesize email: \{kshitij.jindal, aw3645, loiannog\}@nyu.edu}.}
  \thanks{$^{4}$The authors are with the University of Pennsylvania, GRASP Lab, 3330 Walnut Street, 19104, Philadelphia, PA, USA. {\tt\footnotesize email: \{tdinesh, alexzhou\}@seas.upenn.edu}.}

}

\begin{document}

\maketitle

\begin{abstract}
  Mobile manipulators have the potential to revolutionize modern agriculture, logistics and manufacturing.
  In this work, we present the design of a ground-based mobile manipulator for automated structure assembly.
  The proposed system is capable of autonomous localization, grasping, transportation and deployment of construction material in a semi-structured environment.
  Special effort was put into making the system invariant to lighting changes, and not reliant on external positioning systems.
  Therefore, the presented system is self-contained and capable of operating in outdoor and indoor conditions alike.
  Finally, we present means to extend the perceptive radius of the vehicle by using it in cooperation with an autonomous drone, which provides aerial reconnaissance.
  Performance of the proposed system has been evaluated in a series of experiments conducted in real-world conditions.
\end{abstract}

\section{Introduction}
Robotic assembly has become a staple of the manufacturing process over the past decades.
For many years, industrial robots have been purpose-built to perform a singular task with nearly zero versatility or possibility to reconfigure the assembly process.
The status quo is slowly changing with the increasing availability of compact but powerful computational units and new lightweight actuators.
The available processing power allows for extensive use of advanced planning and perception methods, as well as deep learning techniques supported by a widespread availability of datasets \cite{reco}.

\begin{figure}[ht!]
  \centering
  \begin{subfigure}{0.49\columnwidth}
    \includegraphics[width=\textwidth]{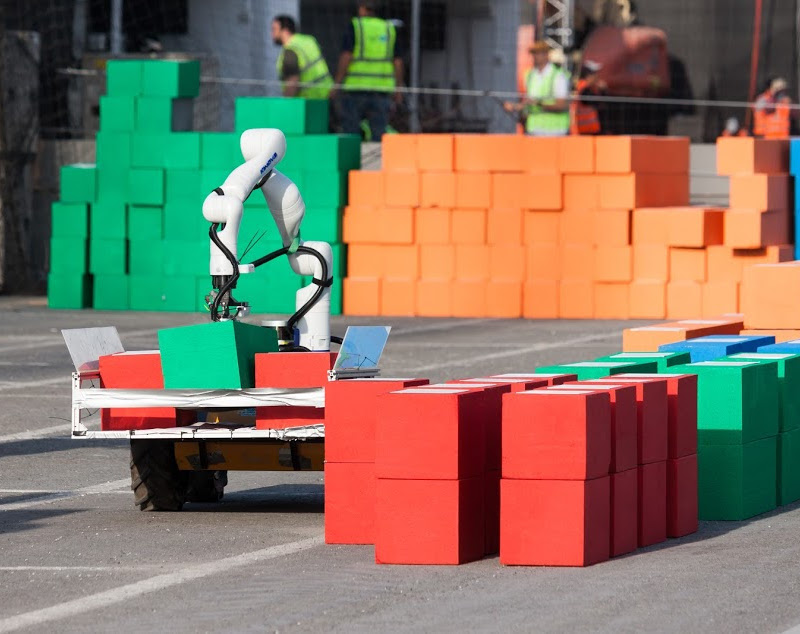}
  \end{subfigure}
  \begin{subfigure}{0.49\columnwidth}
    \includegraphics[width=\textwidth]{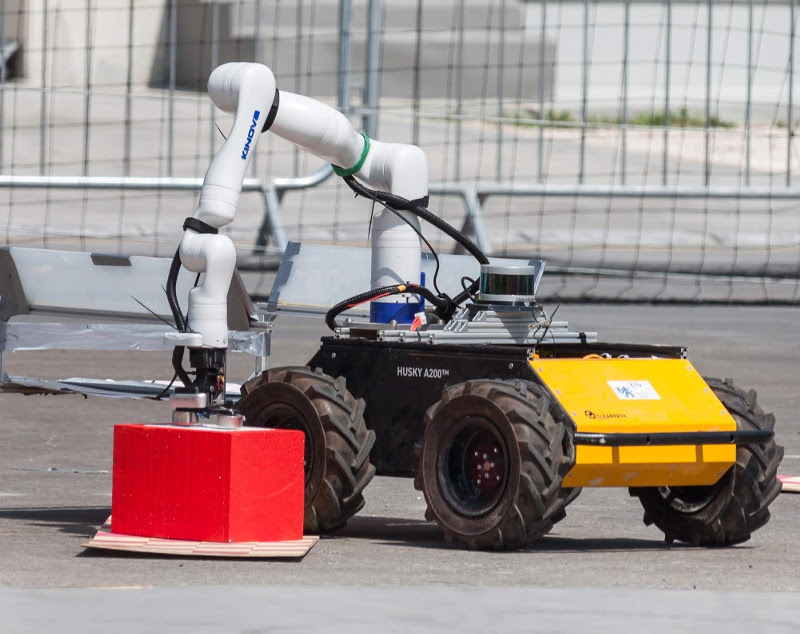}
  \end{subfigure}
  \caption{The described system loading and placing building material during the MBZIRC 2020 contest.}
  \label{fig:promo}
\end{figure}

The growth in e-commerce also led into the development of robots in warehouses, where they are used for automated storage or product retrieval \cite{bogue2016growth}.
Moreover, robots also see increased prevalence outside of the structured environments, e.g., in households and offices \cite{strands,spencer}.
The construction task addressed in this paper takes the logistic aspects of warehouse automation and employs them into a less organized semi-structured environment, as shown in Fig. \ref{fig:promo}.

Despite the undeniable progress made in recent years, common household robots like intelligent vacuum cleaners are still a long way from versatile robotic assistants.
The challenges of designing such system include navigation in a cluttered dynamic environment and interaction with a wide variety of household objects.
Most importantly, the operation has to be reliable and safe, especially since the robot is expected to share its operational space with humans.

Unfortunately, the design requirements often contradict each other.
Excellent dexterity can only be satisfied by a large robotic arm with many actuators, which in turn results in a need for a large and stable mobile base.
However, safe and precise navigation in a cluttered environment is often difficult for bulky robots.
Therefore, present day mobile manipulators appear as clumsy and require human supervision while operating, which somewhat negates the purpose of an automated system.
This becomes even more pronounced when the operation spans over larger time periods \cite{strands,lta}.
The need for human supervision also makes the mobile manipulator economically unviable.

The economical aspects may be offset by considering environments, which are potentially hazardous for humans.
These include search and rescue operations in underground environments \cite{rouvcek2019darpa,petrlik2020robust}, clearing areas of toxic or radioactive debris \cite{amjadi2019cooperative,nawaz2009underwater}, high-rise building maintenance \cite{maintenance} or power line inspection \cite{9213924, 9213967}.

  \subsection{Related work}
The deployment of complex cyber-physical systems and machine learning in industry marks a historic shift in design paradigms, which has been dubbed the fourth industrial revolution, or Industry 4.0 \cite{lu2017industry, lasi2014industry}.

Commercially available mobile manipulators include the KUKA KMR iiwa\footnote{\url{https://www.kuka.com/en-de/products/mobility/mobile-robots/kmr-iiwa}}, Robotnik RB-1\footnote{\url{https://robotnik.eu/products/mobile-manipulators/rb-1}} or the Fetch Robotics Mobile Manipulator \cite{wise2016fetch}.
Despite availability, mobile manipulators remain predominantly a subject of academic research.

A 5 DOF manipulator mounted to an omni-directional mobile platform proposed in \cite{bischoff2011kuka} demonstrated the ability to localize and pick up colored objects in a semi-structured environment.
A dexterity analysis in \cite{chen2018dexterous,domel2017toward} considers the task of picking up an object from a shelf by a mobile manipulator.
In \cite{ohashi2016realization}, manipulation and transportation of a heavy payload is addressed.
Finally in \cite{pavlichenko2018kittingbot}, a robot is shown to fetch various warehouse objects, while also avoiding a human worker, who occupies the same area.

Mobile manipulators will play a crucial role in the transition to Industry 4.0, as they offer an easy way to reconfigure the assembly line by self-reorganization, and provide more options for cooperation with human workers.
In the near future, mobile manipulators may become an integral component of smart factories, workshops and highly modular assembly lines \cite{xu2018industry,petrasch2016process,roblek2016complex,davis2012smart}.\looseness=-1

The field of precision agriculture may also greatly benefit from mobile manipulators.
However, robots working outdoors have to overcome changing weather and lighting conditions \cite{duckett2018agricultural}.
In \cite{bac2014harvesting}, more than 50 agricultural mobile manipulation systems were reviewed.
The results show a significant lack of reliability, if the robots are deployed outside of laboratory conditions.
However, the challenges of reliable robotic perception in the real-world outdoor environment are slowly being overcome, as new sensors and processing methods are developed \cite{ponnambalam2020agri,binch2020context,kusumam20173d}.
Novel approaches also allow for long-term deployment of agricultural robotic systems \cite{pretto2020building}.
A study in \cite{bogue2020fruit} shows, that specializing in one crop type and adjusting the environment for robotic operation significantly improves reliability in real-world conditions.

Aiming to push the boundaries of research, the Mohamed Bin Zayed International Robotics Challenge\footnote{\url{http://mbzirc.com}} (MBZIRC) held in 2017 and in 2020 also focused on mobile manipulation tasks.
The competition attracted attention and participants from renowned research institutions from all over the world.
In 2017, a ground robot was tasked to autonomously locate a valve, pick up a tool of an appropriate size and turn the valve \cite{schwarz2019team}.
It also featured the ``Treasure hunt'' challenge, where a team of up to 3 aerial robots was tasked to locate, grasp and transport magnetic discs to a designated drop-off zone \cite{spurny2019cooperative, baca2019autonomous, loianno2018localization}.
In 2020, a successor to the Treasure hunt was introduced in the ``Brick building'' challenge.
The difficulty was increased significantly by adding objects of varying sizes, replacing the drop-off by precision placement, and requiring both ground and aerial vehicles to be deployed simultaneously \cite{penicka2020mbzirc}.

  \subsection{Contribution}
We present a self-contained robotic system for autonomous localization, grasping, transportation and precise placement of magnetic blocks.
The system design is provided to the community together with the necessary onboard software as open-source \cite{github}.
In contrast to the commercially available mobile manipulators, the proposed system is highly modular and dexterous, while maintaining a compact form-factor.
As was mentioned earlier, these two design requirements often contradict each other.
The proposed system greatly benefits of a unique combination of a compact, yet powerful mobile base, and a very lightweight ($8.2$~kg) manipulator arm with 7 DOF, which is able to carry objects weighing up to $3$~kg.
Another major advantage is the ability to traverse uneven terrain, which is essential for outdoor operation, but is not present with the aforementioned industrial solutions.

We propose a modified variant of the fast segmentation algorithm introduced in \cite{krajnik2014jint} to efficiently extract objects of interest from both depth and RGB image streams.
The system is designed to be invariant to external lighting, and was proven to work in darkness as well as broad daylight.
We also show, how to enhance the limited perception range of a ground-based robot by employing a cooperating aerial vehicle equipped with an onboard camera, to quickly detect visually distinct features of the area.


Finally, we share the experience gained by our participation in the Brick challenge of the MBZIRC 2020.
Out of 19 teams participating in this challenge, only 2 teams managed to complete the task with their ground robots in autonomous mode.
As the winners of the competition, we believe that presenting the complete system to the community will make a valuable contribution.

\section{Problem description}
This work tackles a construction task in an outdoor semi-structured environment.
The area of operation is rectangular, approximately $50$~m~$\times$~$60$~m in size.
It is largely devoid of features and obstacles.
One unmanned ground vehicle (UGV) and three unmanned aerial vehicles (UAV) are deployed in the area simultaneously.
The following objects of interest are located in the area: stacked bricks for the UGV to pick up and a colored pattern to place the bricks to.
The arena also contains stacked bricks and a placement area for the UAVs.
These objects are not of interest to the UGV and therefore will be treated as obstacles.
The objects may be located at an arbitrary position in the arena.
A precise position and orientation of these objects is not known in advance.

\begin{figure}[thpb]
  \centering

  \begin{subfigure}[b]{.78\columnwidth}
    \includegraphics[width=1.0\textwidth, clip, trim=1.7cm 0 0 0]{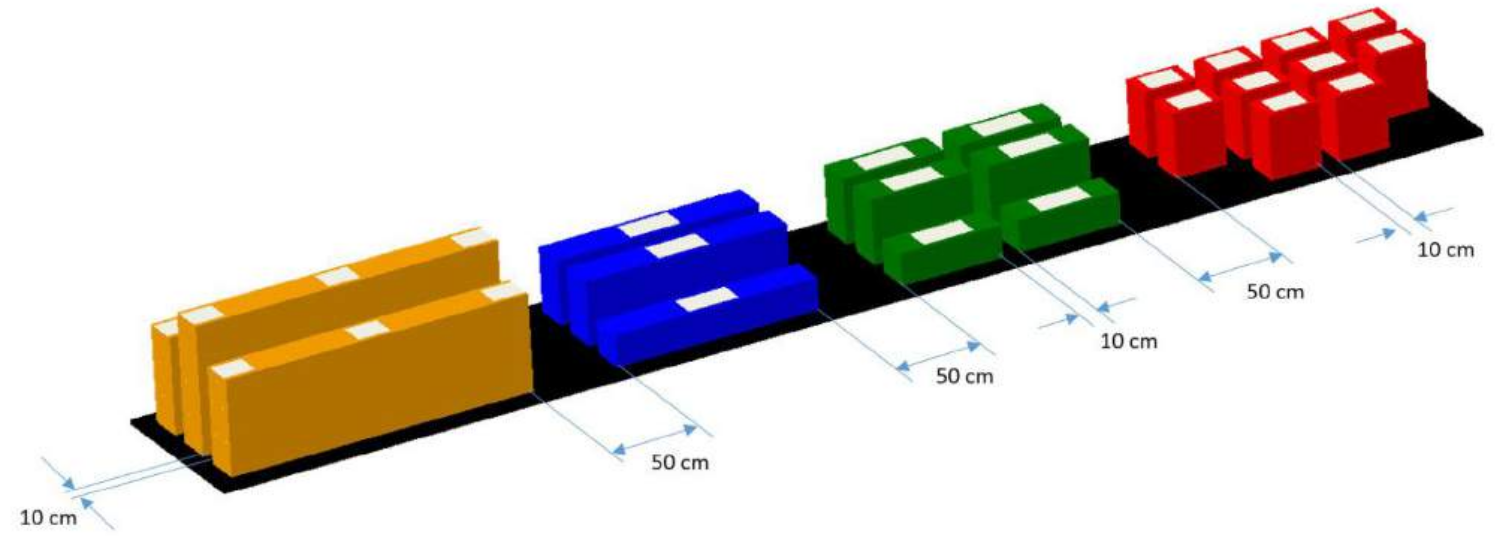}
    \label{fig:brick_pile}
  \end{subfigure}
  \begin{subfigure}[b]{.2\columnwidth}
    \includegraphics[width=1.0\textwidth]{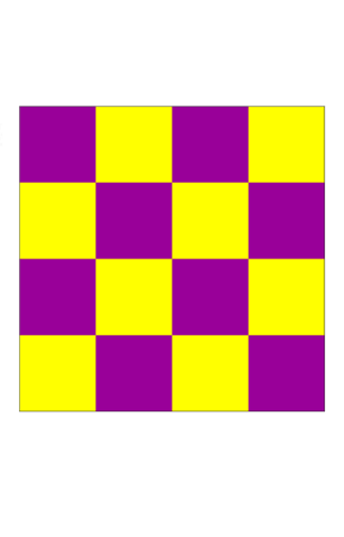}
    \label{fig:build_pattern}
  \end{subfigure}
	\caption{Objects of interest, with which the UGV will interact. The brick pickup area (left) consists of colored bricks stacked in multiple layers. The checkered pattern (right) indicates the deployment zone.}
  \label{fig:objects_of_interest}
\end{figure}

The stacked bricks are arranged in a predefined pattern, as shown in Fig. \ref{fig:objects_of_interest}.
There are four classes of bricks, which differ in size, weight and color.
Each brick features a white ferrous plate attached to the top side.
The placement area consists of two flat segments forming an L-shape.
The segments are covered in a high contrast yellow-magenta checkered pattern, see Fig. \ref{fig:objects_of_interest}, and each segment is $4$~m long, and $0.4$~m wide.
The pattern is laid directly on the ground.

The structures related to UAVs are clearly distinguishable, with the pickup area only consisting of one layer of bricks, and the placement area being placed on an elevated platform with a height of $1.7$~m.

The goal is to use the colored bricks to construct a wall in the area marked by the checkered pattern.
Each layer of the wall consists of exactly 2 orange, 1 blue, 2 green and 4 red bricks. The ordering of the bricks is specified in advance, but changes with each trial.
Each correctly placed brick is awarded by a score gain, with larger bricks being worth more points.
The time limit to complete the challenge is 30 minutes.

\section{Robotic platform}

\begin{figure}[htbp]
  \centering
  \begin{subfigure}{0.49\columnwidth}
    \includegraphics[width=\textwidth]{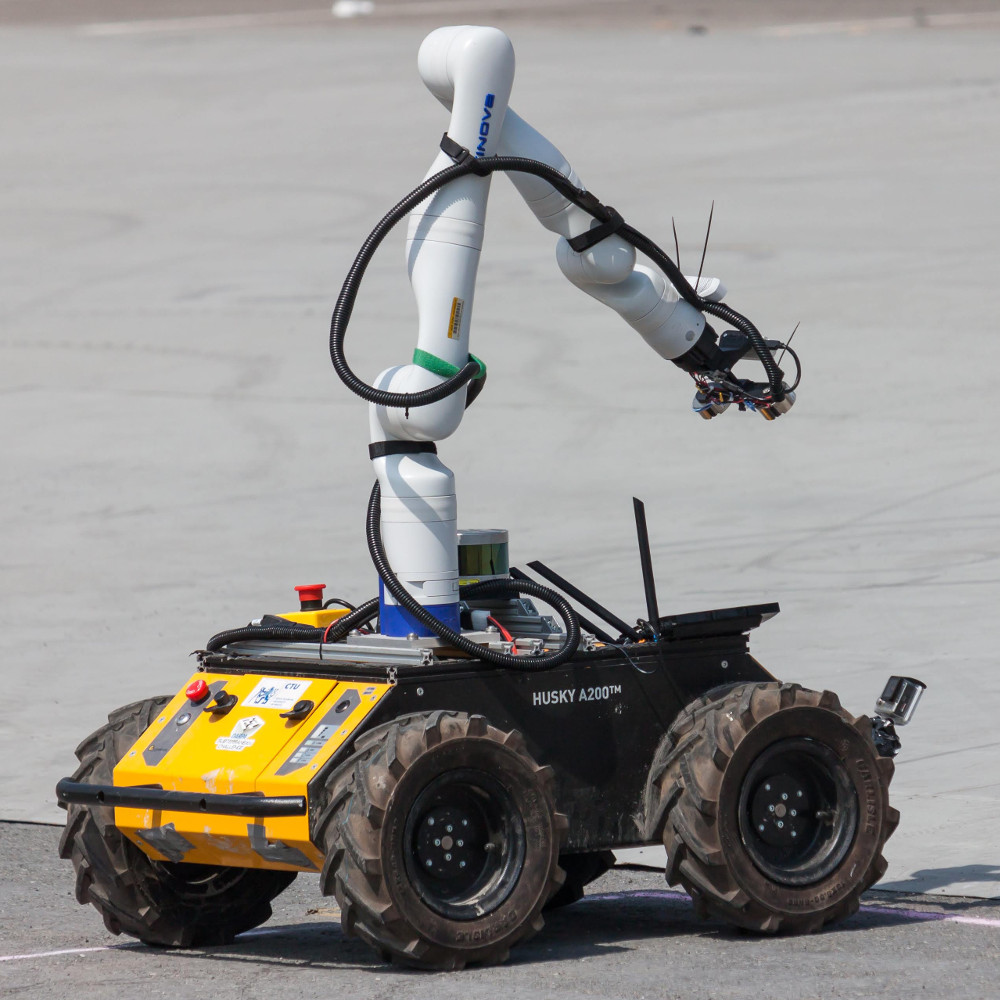}
    \subcaption{Proposed mobile manipulator}
    \label{fig:folded_arm}
  \end{subfigure}
  \begin{subfigure}{0.49\columnwidth}
    \includegraphics[width=\textwidth]{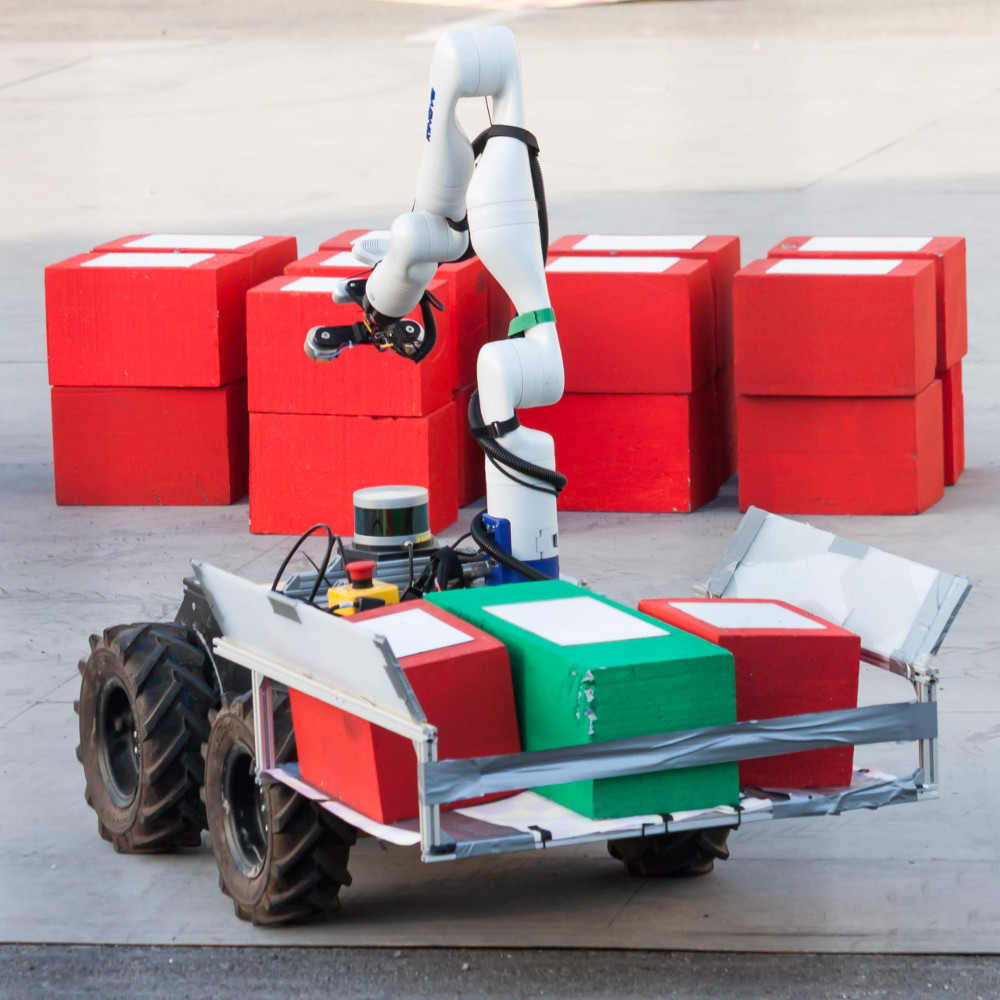}
    \subcaption{Cargo bay attached}
  \end{subfigure}
  \caption{An overview of the proposed robotic system. A Gen3 Kinova manipulator arm is attached to a Clearpath Husky wheeled base. A custom-designed cargo bay is attached to the robot to increase payload capacity and minimize the time spent by traversing the construction site.}
  \label{fig:platform}
\end{figure}

The proposed system is built using mostly off-the-shelf components, which corresponds well with our modular and open design philosophy.
A Clearpath Robotics Husky A200 serves as the mobile base for the system.
An Intel NUCi7, with the operating system Ubuntu 18.04, is used as the main onboard computer.
The Robot Operating System (ROS) serves as a middleware interconnecting the hardware and software parts \cite{quigley2009ros}.
For object manipulation, a 7 DOF Kinova Robotics Gen3 arm is employed.
The arm is mounted to the mobile base as shown in Fig. \ref{fig:platform}.

\begin{figure}[htbp]
  \centering
  \begin{subfigure}{0.49\columnwidth}
    \includegraphics[width=\textwidth]{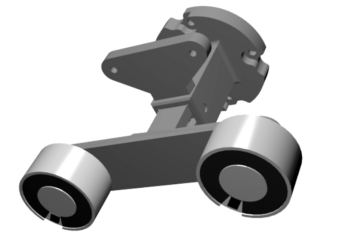}
    \subcaption{CAD design}
  \end{subfigure}
  \begin{subfigure}{0.49\columnwidth}
    \includegraphics[width=\textwidth]{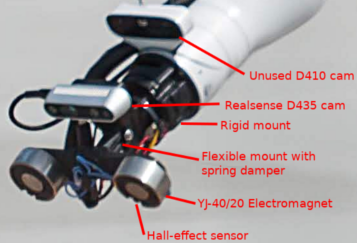}
    \subcaption{3D printed and mounted}
  \end{subfigure}
  \caption{Detail of the custom-designed magnetic gripper for the end-effector.
  The gripper is fitted with two electromagnets and Hall-effect sensors.
  The design also incorporates a mounting point for the Realsense D435 camera, which is used for visual servoing.}
  \label{fig:gripper_detail}
\end{figure}

The arm is equipped with a custom-designed, force compliant end-effector with a magnetic gripper as shown in Fig. \ref{fig:gripper_detail}.
Its design is based on our previous experience with magnetic object grasping by aerial vehicles \cite{spurny2019cooperative, loianno2018localization}.
The gripper features two YJ-40/20 solenoid electromagnets, mounted at a $130$~mm center-to-center pitch.
The magnets are rated to operate at $12$~V, with each magnet providing up to $25$~kg of holding force.
However, the force is strongly affected by the relative distance and orientation of contact the surfaces.
In our setup, the magnets are overcharged with a $24$~V input, which effectively doubles the available grasping force.
On the other hand, it makes the gripper susceptible to overheating when powered on for extended periods of time.
Each magnet is fitted with a Hall-effect sensor, which serves as a proximity detector for ferromagnetic objects.
The sensoric readout and power management of the magnets is handled by an Arduino Nano board, which is connected via USB to the main computer.

To minimize the time spent by transitions between the material loading area and the deployment area, a custom-designed cargo bay is attached to the rear of the robot.
The cargo bay allows the robot to store up to 7 bricks (4 red, 2 green and 1 blue), which are sufficient for construction of an entire section of the structure.
Loading a brick into the cargo bay takes around 30 seconds.
During this time period, the gripper temperature is kept to reasonable levels.

For environment sensing, the robot is fitted with a Velodyne Puck VLP-16 LiDAR\footnote{Light Detection and Ranging}, consisting of a rotating infrared laser range-finder.
The LiDAR provides a $360^{\circ}$ horizontal and a $30^{\circ}$ vertical field of view, with an effective range of around $50$~m.
The sensor is connected to the main computer via Ethernet, and provides measurements at a rate of $10$~Hz in the form of a 3D point cloud.

The manipulator is already fitted with an Intel Realsense D410 RGBD camera integrated into the wrist link.
This camera was intended to provide visual feedback during the brick grasping process.
However, we have experienced a significant delay in the image output, which rendered it impossible to use in a feedback loop.
Therefore, an additional Intel Realsense D435 RGBD camera was mounted to the gripper and connected directly to the onboard computer.
The camera provides an RGB resolution of up to $1920 \times 1080$ pixels, and a depth image resolution of up to $1280 \times 720$ pixels.

\section{Control architecture}
The following sections summarize the key software components used to successfully complete the task in autonomous mode.
As mentioned before, the environment layout is unknown prior to the deployment.
The robot therefore has to explore the area and localize the objects of interest.
Despite using an array of powerful onboard sensors, the object detection range of the robot is still quite limited due to the finite resolution of the sensors and the size of the objects.
For this reason, we concluded that a full exploration of the area would not be feasible given the limited time budget.
On the other hand, specifying navigational waypoints in an unknown environment is not possible.

We approach the problem using a ``guided autonomy'' concept \cite{gray2018architecture,dellin2016guided,hebert2015supervised}, which aims to strike the balance between exhaustive exploration and going directly to a well-defined target.
The system is able to investigate coarsely defined areas of interest, which may be provided on-the-fly by a human operator or by other cooperating robots.
Limiting the search space allows the robot to conserve some of the limited time budget, and if no objects of interest are found within the prioritized areas, the full exploration is resumed.



\section{Arm manipulation}
The arm manipulation is controlled by the Kinova Kortex API and a ROS driver provided by the manufacturer\footnote{\url{https://github.com/Kinovarobotics/ros_kortex}}.
We have designed a custom wrapper for the API called Kortex Control Manager (KCM), which integrates the API into the control pipeline.

\begin{figure}[htbp]
  \centering
  \includegraphics[width=0.8\columnwidth]{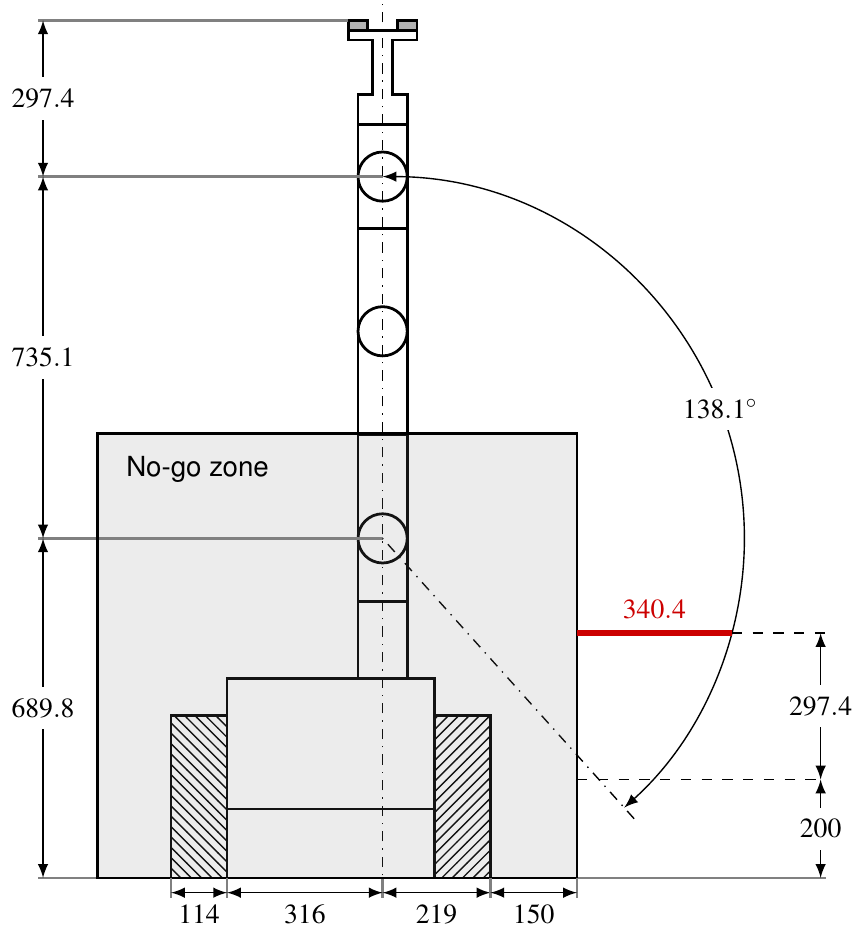}
  \caption{Grasping reach of the proposed mobile manipulator with the 7 DOF Kinova Gen3 manipulator. The arm is equipped with the custom end-effector and a magnetic gripper. If a brick center is located within the range highlighted in red, it can be reliably grasped. The dimensions are shown in millimeters unless stated otherwise.}
  \label{fig:working_envelope}
\end{figure}

The KCM includes a library of short pre-programmed actions, which are chained to form more complex tasks, such as the brick grasping and loading.
In addition to self-collision avoidance, which is integrated into the API, the KCM also expands this by enforcing an extended no-go zone over the UGV's frame. The no-go zone is further expanded by half a brick size to prevent collisions with the grasped brick.
These safety features are necessary, but significantly reduce the operational space of the arm, as shown in Fig. \ref{fig:working_envelope}.
The limited reach of the arm therefore needs to be compensated by the mobile base.

The KCM offers position control of the manipulator in joint space, and velocity control of the end-effector in Cartesian space.
In the position control mode we utilize the MoveIt! planning interface \cite{chitta2012moveit} to interpolate the trajectory between current and desired position.
The trajectory is sampled with a time step of $1$~ms and the corresponding joint commands are sent to the actuators via the Kortex API.

The KCM also monitors the data returned by the integrated sensors, which measure position, velocity and torque in each of the joints at a rate of $100$~Hz.
The data is analyzed, and the motion is terminated if excessive torque or no-go zone violation is detected.

\section{Navigation and localization}
The UGV uses a default ROS navigation stack, as implemented by the STRANDS project \cite{strands}.
The main source of positional data for navigation is the onboard LiDAR.
During the development, we have discovered that the dynamically changing area of a construction site and a non-flat ground plane introduce a lot of false-positive measurements to the localization stack.

By experimental evaluation, it was found that the localization performance can be significantly improved by focusing on the upper layers of the point cloud.
Therefore, the LiDAR scan is separated into two horizontal slices, with the upper slice being used for robot localization, and the lower slice for obstacle avoidance.
For this particular application, the slicing threshold was empirically set to $1.5$~m above the sensor.

Prior to the deployment, a high-resolution scan of the surrounding environment is taken by a Leica BLK360 3D scanner.
The point cloud obtained by the 3D scanner is used to augment the measurements taken by the onboard LiDAR.
The augmentation contributes significantly to the quality of localization, especially when the UGV moves near the area center, where the feature density is the lowest.

The pose of the UGV is estimated using the Adaptive Monte Carlo Localization (AMCL)\footnote{\url{http://wiki.ros.org/amcl}}.
The AMCL essentially attempts to correct the drift in measured wheel odometry by matching onboard laser scans to the area map \cite{fox1999monte}.
However, the odometry estimator of the Husky gets distorted by shifting the center of mass of the vehicle, which occurs with every manipulator movement and with every brick added into the cargo bay.
We have managed to solve this issue by exhaustively tuning the parameters of the odometry estimator for various vehicle and payload configurations.

The ROS move\_base\footnote{\url{http://wiki.ros.org/move_base}} package is employed for path planning.
Several adaptations of the code were made to allow the UGV to dynamically change the precision of path following.
During the exploration phase, the precision is relaxed, as the focus is on covering large area in the shortest amount of time.
On the contrary, while approaching the stacked bricks, high precision is strongly required, since the area reachable by the manipulator is limited.
In the high precision mode, the UGV performs a series of ``parallel parking'' manoeuvres to reach the desired position with very tight margins.

\begin{figure}[thpb]
  \vspace{0.3em}
  \centering
  \includegraphics[width=0.75\columnwidth]{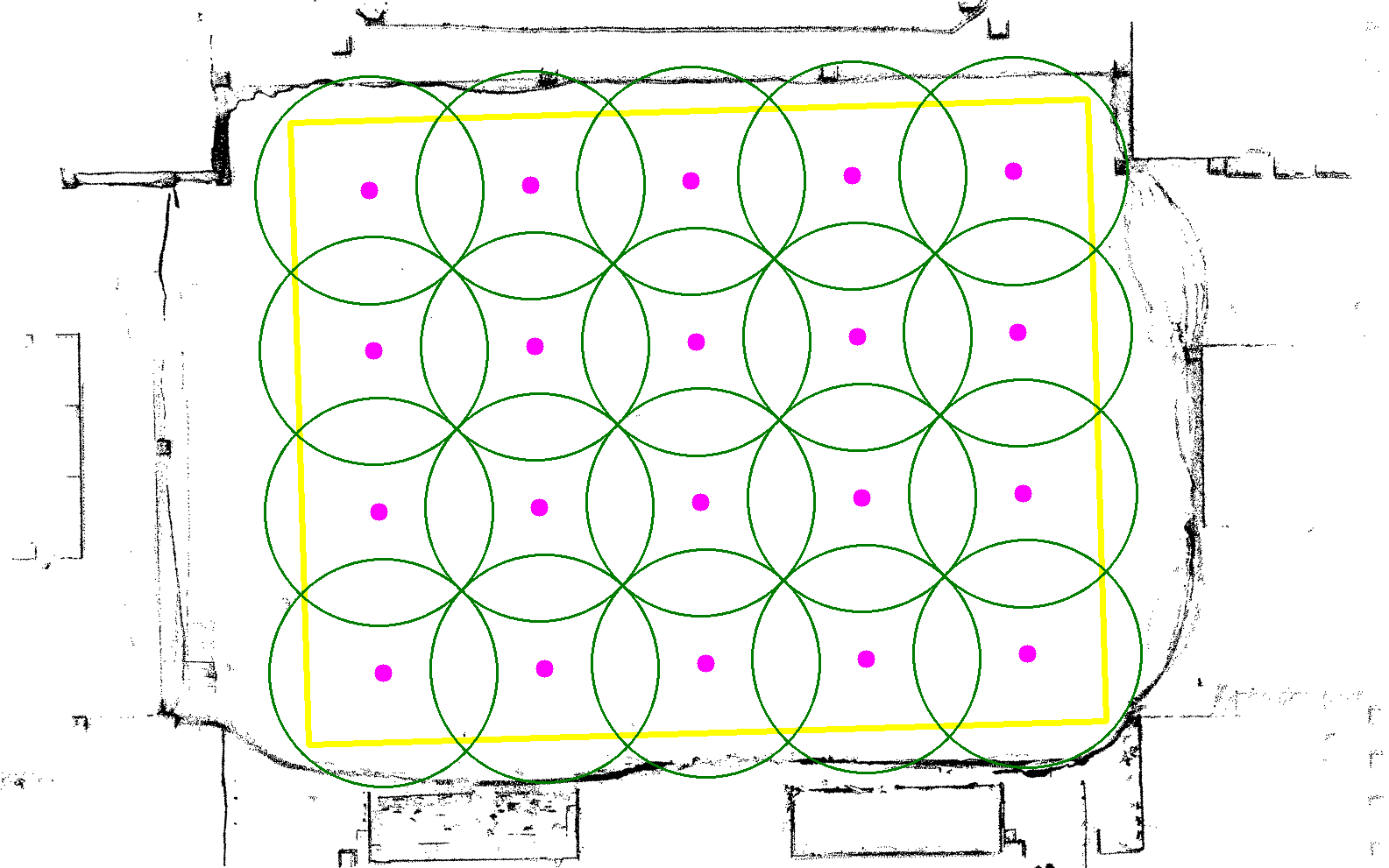}
  \caption{Visualization of the full coverage exploration strategy. The high resolution map obtained by the 3D scanner is outlined in black. The yellow frame determines the operational area, the navigational waypoints are shown in purple. The green circles centered around each waypoint represent the maximal perception range of the system. The UGV then follows a zig-zag pattern in order to visit all the waypoints.\looseness=-1}
  \label{fig:exploration}
\end{figure}

Without any additional information, the UGV conducts a full exploration of the area by generating a grid of waypoints to achieve a full sensoric coverage of the area, see Fig. \ref{fig:exploration}.
An early termination of the procedure is triggered, if all objects of interest are located.

\section{Automated construction}
  \subsection{Inventory management}
The proposed design features a cargo bay, which can hold up to 7 bricks.
However, in order to preserve a compact form factor, the bricks in the cargo bay are stacked into three layers.
This renders some bricks unreachable, if other bricks are loaded on top of them.

We have developed an inventory management system, which monitors the reachability of individual bricks.
The inventory manager also processes the blueprint for the desired wall assembly, and determines which bricks to pick up, and where to place them.
This module also allows the UGV to dynamically change the building strategy.
This feature is especially useful when the UGV encounters an unexpected issue, such as an inability to grasp some of the bricks.
In such case, a new strategy is to immediately navigate to the building pattern and unload all the remaining cargo.


  \subsection{Brick stack detection}
During the exploration phase, the LiDAR is used for detection of the stacked bricks.
The Velodyne VLP-16 has a $30^{\circ}$ vertical field of view divided into 16 evenly spaced scan layers.
To reliably distinguish a brick in the point cloud, at least two scan layers have to hit it.
The number of rays hitting a brick can be determined as
\begin{equation}
  \phantom{,.}N = \frac{\mathrm{arccos}\left(1-\frac{a^2}{2b^2}\right)}{\alpha}\phantom{.},
\end{equation}
where $a$ is the brick height, $b$ is the distance between the sensor and the point (ray length) and $\alpha$ is the angular pitch of two neighboring scan layers.
The maximal detection distance $d$ is then estimated as
\begin{equation}
  \phantom{..}d = \frac{\frac{a}{4}}{\mathrm{tan}\left(\frac{\alpha}{2}\right)}\phantom{.}.
\end{equation}
For a single brick with a height of $0.2$~m, the estimated detection range is approximately $3.055$~m.
The formula assumes an idealized scenario, in which the rays and one brick side form an isosceles triangle.
In practice, the range would be even shorter.
Fortunately, the apriori knowledge of the stack layout can be leveraged to increase the range, since all brick types are arranged in at least two layers.

The point cloud is processed using the Iterative End Point Fit (IEPF) algorithm \cite{ramer1972iterative}, which extracts line segments from the measurement.
All segments of length matching one of the brick classes are considered brick candidates.
In the real-world, this approach alone is not sufficient, as the structure of the environment causes generation of false-positive candidates.

\begin{figure}[htbp]
  \centering
  \includegraphics[width=0.7\columnwidth]{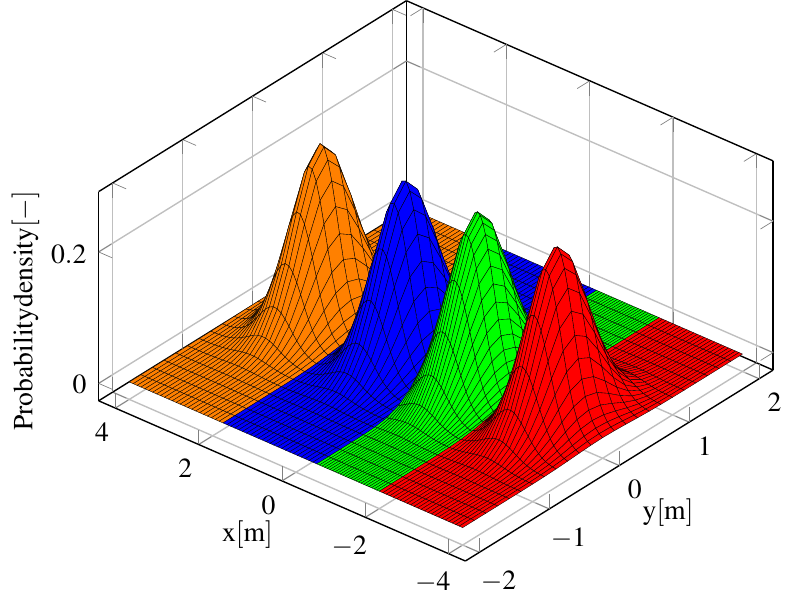}
  \caption{Probability density of the brick class position in the pickup area. This model is used by the brick classifier to estimate the position and orientation of the pickup area.}
  \label{fig:classifier_model}
\end{figure}

To filter out the erroneous data, an Expectation Maximization (EM) algorithm is employed.
The algorithm uses the following probabilistic model:
\begin{equation}
  \phantom{,.}P(\vec{x}_m) \sim  \mathcal{N}(\vec{x}_m, \, \vec{\mu} + k_m\vec{v}, \, \bm{\Sigma_m})\phantom{.},
  \label{eq:prob}
\end{equation}
where $m$ is the brick class index, $\mathcal{N}$ is the Gaussian distribution, $\vec{\mu}$ is the mean of the model, $\bm{\Sigma_m}$ is the covariance matrix of a brick class, $k_m$ is the scalar multiplier unique for each class and $\vec{v} = (\cos\phi, \sin\phi)^T$ is a direction vector representing orientation.
The probability density of individual brick classes is visualized in Fig. \ref{fig:classifier_model}.
The model provides the most likely estimate of the parameters $\vec{\mu}$ and $\phi$.
The parameters correspond with the position of the brick stack center, and the orientation of the stack.
The results are then used to generate a navigational waypoint near the red bricks, so that the bricks are on the right hand side of the robot.

After approaching the red bricks, a line-fitting algorithm based on the Random Sampling Consensus (RANSAC \cite{fischler1981random}) is applied to the point cloud.
The UGV performs a parallel parking action to align its heading with the major axis of the brick stack, which is estimated by the RANSAC algorithm.
After aligning, the manipulator is extended into a predefined position to the right hand side of the Husky.
In this position, the wrist-mounted camera and the gripper point directly towards the ground.

  \subsection{Brick detection, grasping and loading}
The brick detection pipeline is built on the fast image segmentation proposed in \cite{krajnik2014jint}.
The algorithm has been successfully deployed in the Treasure hunt challenge of MBZIRC 2017 \cite{vstvepan2019vision,spurny2019cooperative, loianno2018localization} for detection of circular objects.
The original algorithm works with data provided by an RGB camera.
For the brick detection, we opted to use a depth camera, which proved to be more resilient to illumination changes.
Therefore, the algorithm was adapted for depth image processing, which enabled the robot to operate in direct sunlight and also at night.\looseness=-1

The algorithm factors in the position of the arm to estimate the height of the camera above the ground plane.
The depth information contained within each pixel is transformed into height above the estimated ground level.
The algorithm begins by discarding all data below the height of $0.1$~m.
Then, a flood-fill algorithm described in \cite{krajnik2014jint} is initiated to extract brick candidate segments.
In addition to the segments, the algorithm also obtains their centers, sizes, eccentricities and distances from the camera.
Finally, the method determines, whether all edges of the segment are visible by the camera.

The processing pipeline then attempts to assign a brick class to the fully visible segments.
It starts by locating a pixel with the highest distance from the center, which we denote as $c_0$.
Then, a pixel $c_1$ is found by searching for the maximal distance from $c_0$.
The third corner $c_2$ is found by searching for a maximal distance from both $c_0, c_1$.
The final corner $c_3$ is located by maximizing the sum of distances from the other three corners.
Using the coordinates of the corner pixels and the depth information, the segment is transformed from camera coordinates into 3D and the length and width of the object are calculated.
Since the brick classes are clearly distinguishable by dimensions only, RGB processing may be omitted.
Therefore, the system does not require color calibration to the current lighting conditions.

\begin{figure}[htbp]
  \centering
  \begin{subfigure}{0.32\columnwidth}
    \includegraphics[width=\textwidth]{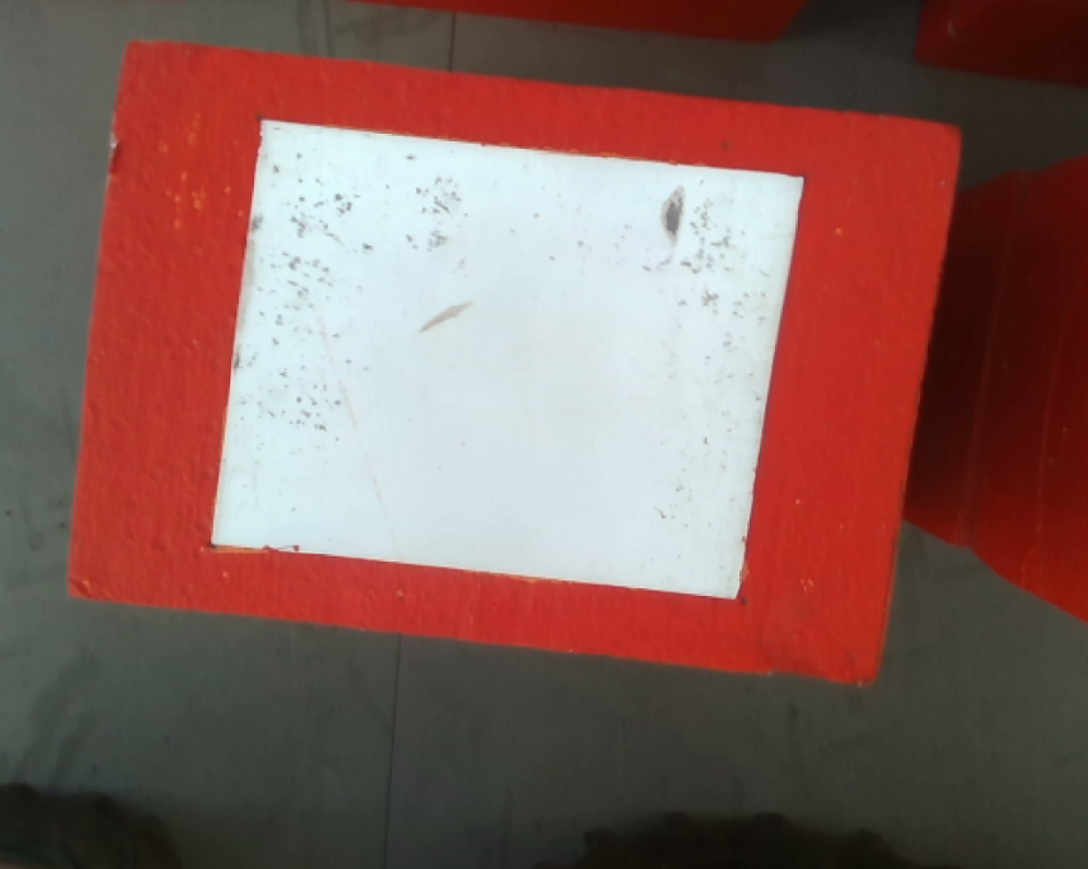}
    \subcaption{RGB image}
  \end{subfigure}
  \begin{subfigure}{0.32\columnwidth}
    \includegraphics[width=\textwidth]{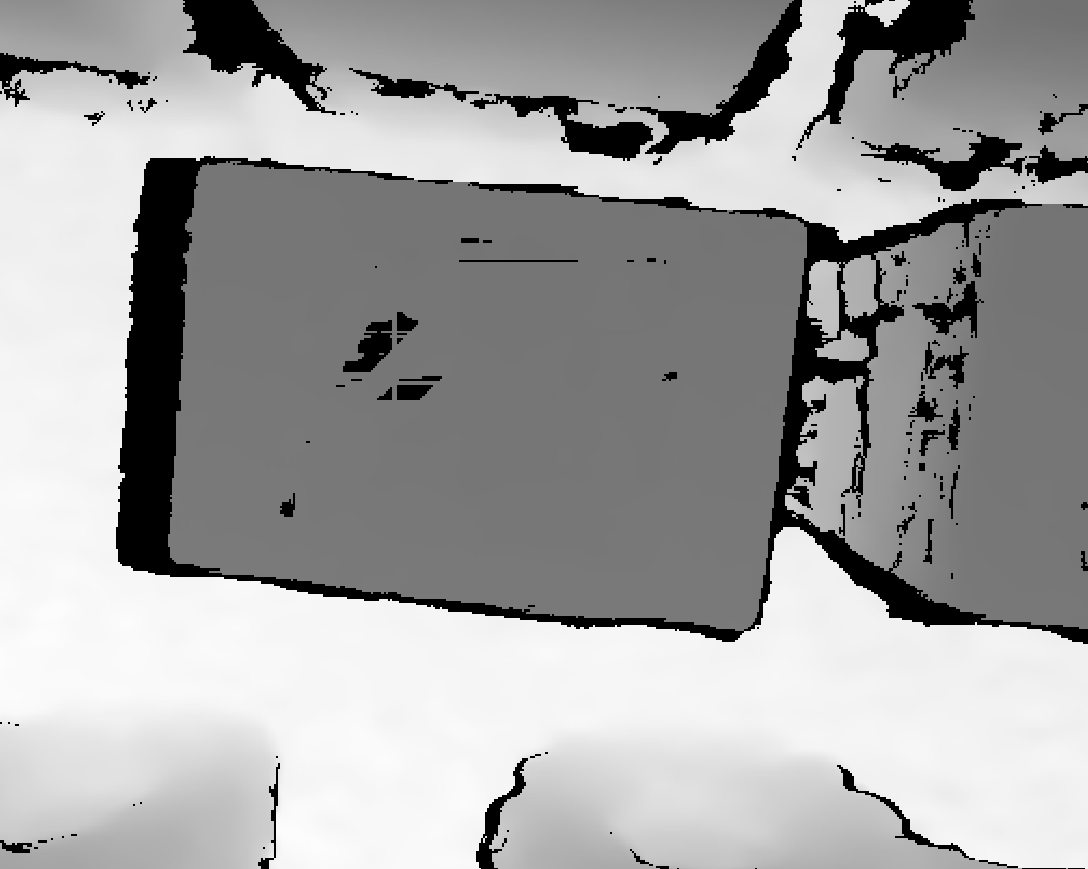}
    \subcaption{Raw depth image}
  \end{subfigure}
  \begin{subfigure}{0.32\columnwidth}
    \includegraphics[width=\textwidth]{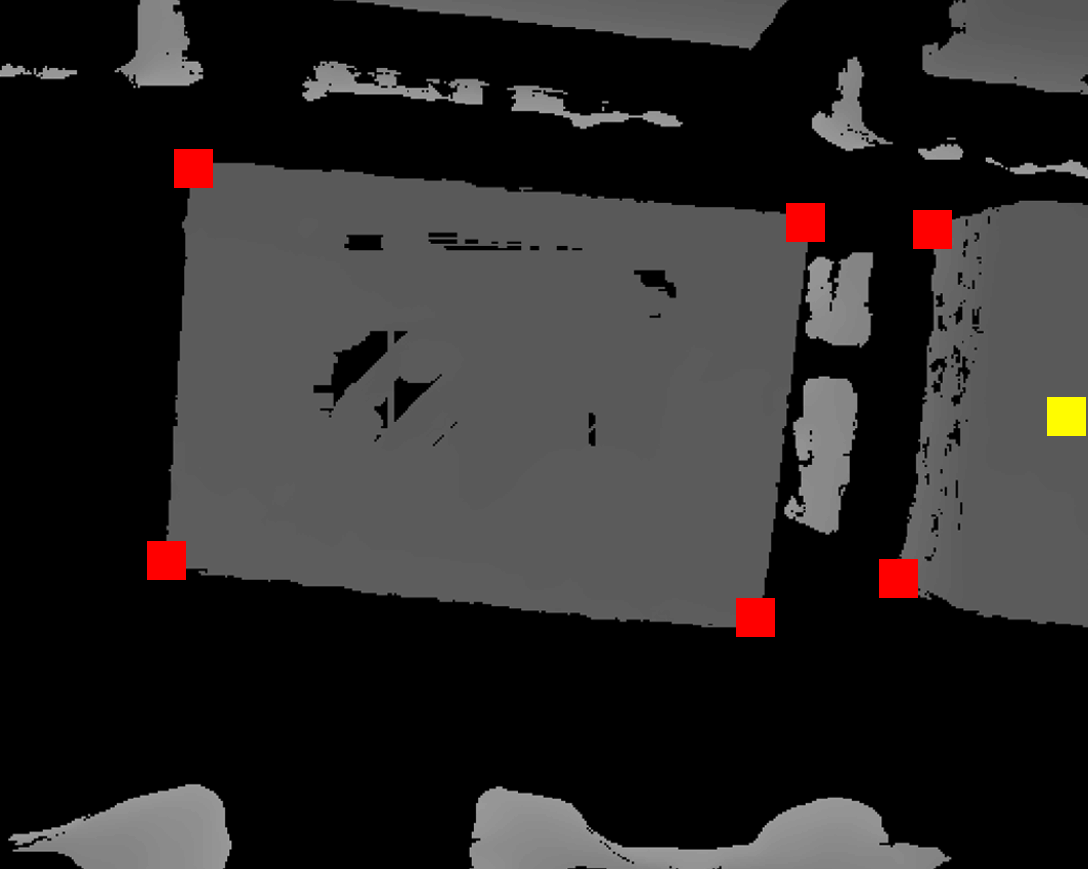}
    \subcaption{Processed depth}
  \end{subfigure}
  \caption{Visual servoing based on the wrist-mounted RGBD camera. The RGB image is only shown for comparison. Due to the tight packing of the bricks, multiple segments are visible at the same time. The segmented depth image is shown on the right, with detected corners marked in red, and the current alignment target in yellow. Note the robot wheels at the bottom of the image.}
  \label{fig:visual_servoing}
\end{figure}

The processed depth image (see Fig.~\ref{fig:visual_servoing}) is used for visual servoing of both the mobile base and the manipulator.
If the image contains the brick class requested by the inventory manager, the segment position is transformed into the coordinate system of the gripper, and the arm attempts to align with its center.
If multiple segments fit the request, the system prioritizes the segment closest to the lower right corner of the image.
If the segment center is unreachable by the arm, the mobile base starts to perform small turns to adjust its position.
The motion is stopped, once the brick center is located directly below the gripper with a $\pm8$~cm tolerance.
Afterwards, the gripper is aligned with the brick center (tolerance $\pm2$~cm) and the arm begins to descend.\looseness=-1

During the descent, lateral adjustments to the gripper position are made as long as the brick is fully visible.
Afterwards, the magnets are powered on, and the arm descends straight down until the Hall-effect sensors report a successful grasp.
Grasping failure is detected by either reaching lower than expected (brick missed completely), or by detecting an excessive torque on the arm joints (ferromagnetic plate missed).
In case of failure, the robot will attempt the grasping process again.
After two failed attempts, the brick is marked invalid, and the UGV moves on to the next one.

Once the brick is grasped, the inventory manager assigns it a position in the cargo bay.
A series of predefined actions ensures that the arm stores the brick in the assigned inventory slot.
Based on the brick class and the current payload status, additional actions are added into the loading sequence to avoid potential collisions.

  \subsection{Build pattern detection}
The checkered building pattern is impossible to detect by the LiDAR, as it blends in with the ground plane.
Therefore we use the RGB portion of the data provided by the arm-mounted RGBD camera.
For the pattern detection, the fast segmentation algorithm based on \cite{krajnik2014jint,vstvepan2019vision} is used.
The algorithm uses a 3D RGB lookup grid, to classify individual pixels as background, object or unknown.
The method enables pixel-wise classification in constant time, which is achieved by pre-computing the lookup grid during a calibration process.

In the case of the high-contrast pattern, the grid is initialized using a Gaussian mixture to model the desired color in the HSV color space.
The RGB image is then searched for the desired color.
Once a satisfactory pixel is found, the flood-fill algorithm is initiated to extract the segment to which the pixel belongs.
In the default state, the algorithm would only find individual squares in the checkered pattern, which would require costly post-processing to merge the objects back together.
Instead, we propose a modified flood-fill algorithm capable of bypassing small discontinuities in the segments.

The algorithm, as presented in \cite{vstvepan2019vision}, naturally obtains the number of colors surrounding each pixel in the segment, allowing for an easy corner detection.
For each corner, the searched neighborhood is increased to 5 pixels.
If a new pixel of the desired color is found within this neighborhood, it is added to the original segment.
This way, half of the entire checkered pattern may be extracted as a single segment.
Comparing the dimensions of the segment with the apriori knowledge of the build area allows for effective filtration of false positive candidates.

\begin{figure}[htbp]
  \vspace{0.4em}
  \centering
  \begin{subfigure}{0.49\columnwidth}
    \includegraphics[width=\textwidth]{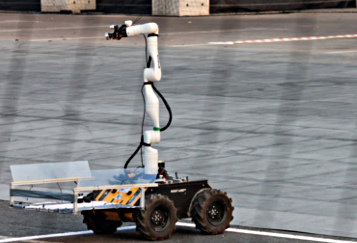}
    \subcaption{Long range pattern search}
  \end{subfigure}
  \begin{subfigure}{0.49\columnwidth}
    \includegraphics[width=\textwidth]{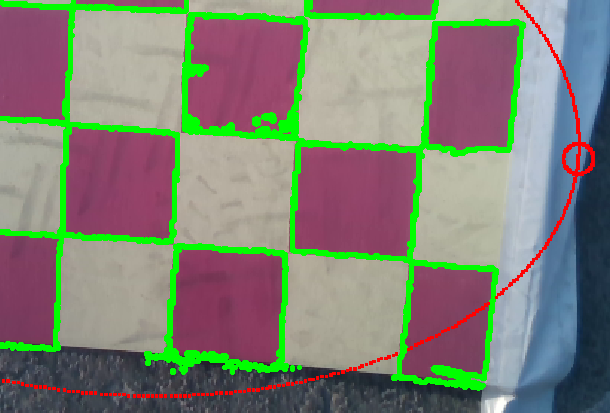}
    \subcaption{Precise pattern alignment}
  \end{subfigure}
  \caption{While searching for the pattern, the UGV extends the arm upwards to provide the camera an elevated view point (left). Doing so requires the vehicle to stop moving due to a large shift in the center of gravity. Once the building area is located, the UGV uses visual servoing to precisely align the mobile base with the pattern (right). Green highlights detected segment edges. The proposed segmentation algorithm enables extraction of all the magenta squares as one object in a single pass, without the need for post-processing.}
	\label{fig:exploarm}
\end{figure}

During the pattern search, the UGV fully extends the arm, placing the camera into an elevated position, see Fig. \ref{fig:exploarm}.
The camera is then rotated to fully scan the surroundings of the robot.
This process requires the UGV to stop moving, as the center of mass is greatly offset by the arm.
Due to the camera resolution and density of the pattern, detection range is limited to approximately $10$~m.
Achieving a full area coverage would require performing the scan at 20 different locations, see Fig.~\ref{fig:exploration}, which is not feasible within the given time limit.

Since the UGV shares the construction area with a team of aerial vehicles, we opted to use a cooperative approach to detect the building pattern.
The UAVs are equipped with a downward facing RGBD camera, and are able to run the pattern detection pipeline on board.
Due to the orientation of the camera, the UAV does not have to stop moving, and the detection may run continuously while in motion.
The detections are reported to the UGV over a WiFi network.
We employ the NimbroNetwork \cite{nimbro_git}, which allows ROS message sharing between multiple independent robotic systems.

Once the approximate position of the build pattern is reached, the arm is folded so that the camera aims forward, as shown in Fig. \ref{fig:folded_arm}.
The UGV then drives in an outward spiral until the pattern appears in the image.
Finally, the arm is extended to the right hand side, and the UGV uses a similar visual servoing approach to align itself with the pattern, as it does with the brick stack, see Fig.~\ref{fig:exploarm}.
The bricks are then deployed in an order determined by the inventory manager.

During the MBZIRC 2020 finals, the UAV assistance was not used due to the fear of overloading the communication channel and disrupting the mutual coordination of the UAVs.
Instead, the pattern position was estimated from images taken during the previous contest rounds, and the cooperation was only emulated.

\section{Experiments}
Early experimental evaluation was performed at the tennis courts of the New York University campus in Abu Dhabi.
In here, the system was repeatably able to locate, load and deploy the bricks in a specified order.
The success rate of brick loading was around $75\%$, with the grasping accuracy reaching $\pm3$~cm.
In the remaining $25\%$ of cases, the system was not able to fill the cargo bay with all seven bricks.
However, in case of a brick loading failure, the system was still able to autonomously interrupt the loading procedure, navigate to the checkered pattern, and unload the contents of the cargo bay.

During the experiments, we measured that navigating to the stacked bricks takes up to 2 minutes, assuming the stack of four orange bricks is visible from almost any starting position.
The precise alignment and loading all 7 bricks into the cargo bay takes nearly 10 minutes.
Without the UAV assistance, the building pattern search is by far the most time consuming operation.
To acquire the image from an elevated position, the robot has to stop, extend the arm upward, look around with the camera and then fold the arm back.
This procedure alone requires 15 minutes on average, which only leaves 3 minutes for the brick deployment.

Increasing the manipulator speed was not possible, as higher accelerations might cause the brick to be dropped.
With the time budget in mind, we also prepared the emergency protocol, which focused on reliably placing a single brick on the first run.
This would reduce the loading and deployment times significantly, leaving enough margin for difficult arena configurations requiring prolonged exploration.

\begin{figure}[htbp]
  \centering
  \begin{subfigure}{0.32\columnwidth}
    \includegraphics[width=\textwidth]{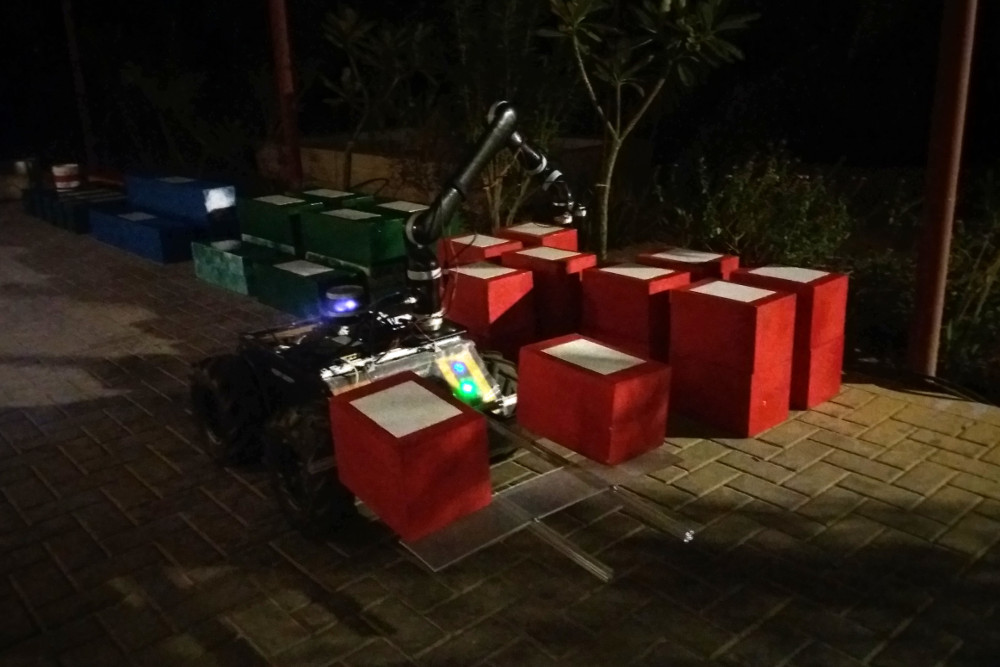}
    \vspace{-1.7em}
    \subcaption{}
  \end{subfigure}
  \begin{subfigure}{0.32\columnwidth}
    \includegraphics[width=\textwidth]{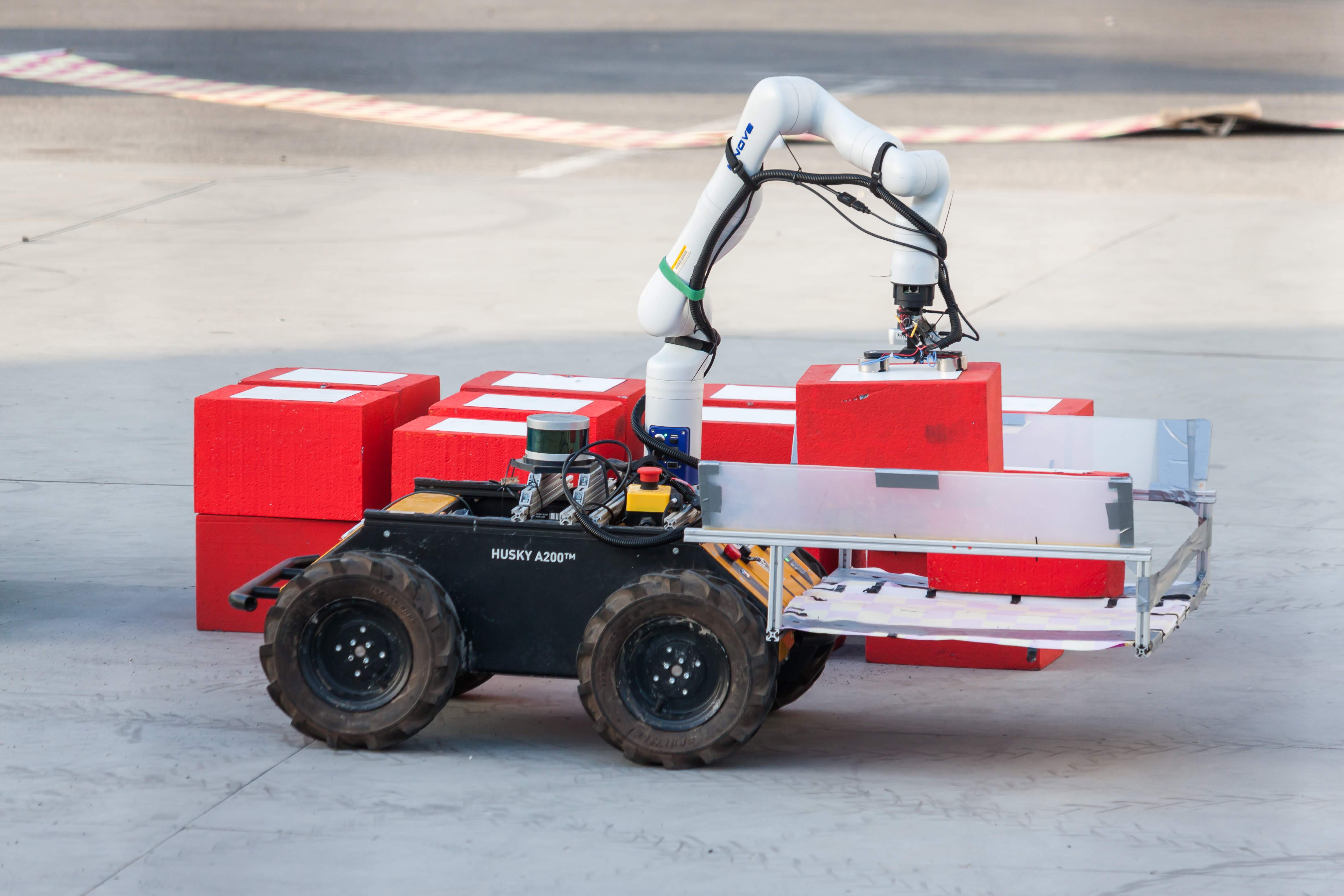}
    \vspace{-1.7em}
    \subcaption{}
  \end{subfigure}
  \begin{subfigure}{0.32\columnwidth}
    \includegraphics[width=\textwidth]{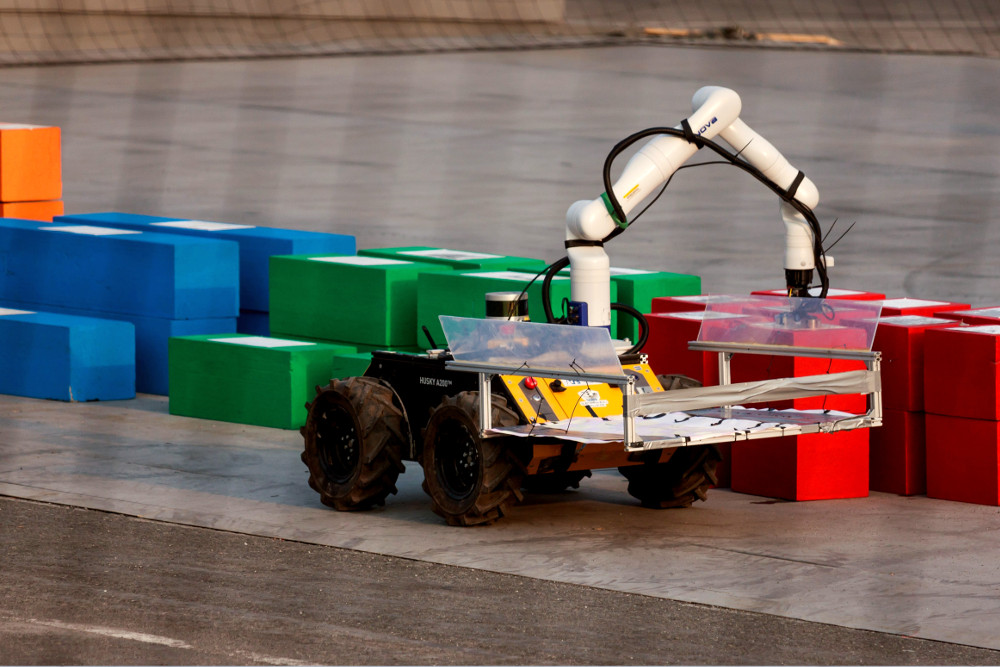}
    \vspace{-1.7em}
    \subcaption{}
  \end{subfigure}
  \vspace{-0.5em}
  \caption{Testing of the grasping procedure in various kinds of lighting conditions. By using the depth image rather than RGB, the grasping can be performed in a repeatable manner at night (a), in direct sunlight (b) or in strong lateral illumination at sunset (c).}
  \label{fig:grasping_illumination}
\end{figure}

The grasping procedure was extensively tested under various lighting conditions to ensure robust performance at any time of day, as shown in Fig. \ref{fig:grasping_illumination}.
This feature greatly enhances the economic feasibility of the proposed system as a robotic construction worker, since night-time construction operations fall into the ``potentially hazardous for humans'' category.

\begin{figure}[htbp]
  \centering
  \begin{subfigure}{0.49\columnwidth}
    \includegraphics[width=\textwidth]{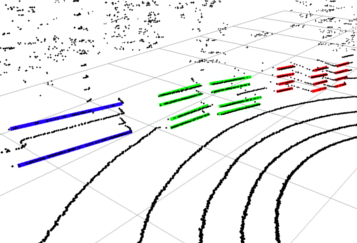}
    \subcaption{Extracted line segments}
  \end{subfigure}
  \begin{subfigure}{0.49\columnwidth}
    \includegraphics[width=\textwidth]{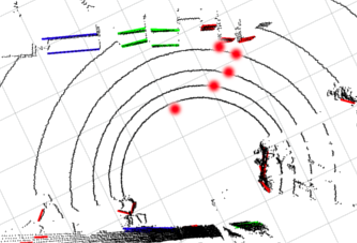}
    \subcaption{Red pile center estimation}
  \end{subfigure}
  \caption{Detection of stacked brick in the LiDAR data. The left image showcases the line segments as extracted and classified by the IEPF algorithm. The right image shows the five initial iterations of the EM algorithm. The red points represent the estimated position of the red pile center.}
  \label{fig:experiments_pointcloud}
\end{figure}

The performance of the brick extraction and classification in the LiDAR data was evaluated offline using a dataset obtained during the first competition rehearsal.
In Table \ref{tab:iepf_classification}, we show a success rate of brick classification by the IEPF algorithm without any additional filtration or odometry fusion.
The blue bricks are the least populous in the pickup area, and therefore are the most susceptible to erroneous detections.
The accuracy is greatly improved by filtering the false positive measurements and by employing the EM algorithm, which incorporates the a priori knowledge of the brick pile layout.
Snapshots of the point cloud processing are shown in Fig. \ref{fig:experiments_pointcloud}.

\begin{table}[htbp]
  \centering
  \begin{tabular}{l|r|r|r}
	  Brick color & Correct & Incorrect & Success rate [\%]\\
	  \hline\rule{0pt}{1.1\normalbaselineskip}
    Red & 174 & 70 & 71 \\
    Green & 57 & 19 & 75 \\
    Blue & 24 & 63 & 28 \\
    Orange & 32 & 11 & 62 \\
  \end{tabular}
  \caption{Accuracy of brick candidates generation by the IEPF algorithm. The evaluation was performed offline using data collected during the first rehearsal in the competition arena.}
  \label{tab:iepf_classification}
\end{table}

The final evaluation took place during the MBZIRC 2020 Brick challenge finals.
During the contest, the brick pickup area was placed on the inclined ramp, which posed additional challenges for the participants.
With no time to reconfigure the system to this new environment, we opted to engage the emergency protocol, which was optimized for reliability and ensured task completion with at least one brick.
The successful brick placement during the contest finals is shown in Fig. \ref{fig:promo}.

\vspace{-0.7em}
\section{Conclusion}
We have presented an autonomous robotic system for localization, grasping, transportation and precise deployment of construction blocks.
The system is capable of autonomous operation in challenging lighting conditions on uneven terrain.
The positioning and navigation pipeline does not rely on external systems, such as the GPS.
The system may therefore be deployed in an indoor environment without any adjustments, and even perform the complicated transition from an outdoor to an indoor environment.
The presented computer vision algorithms, which are application independent, offer fast and reliable segmentation of both depth and RGB images.

We provide all the software to the community as open-source \cite{github}.
The presented system was successfully deployed during the MBZIRC 2020 Brick challenge, which simulated a possible application of mobile manipulators in a construction task.
As the winners of the contest, we hope this work will inspire further development and facilitate the mobile manipulation endeavors of other research groups.
\vspace{-0.5em}

\bibliographystyle{IEEEtran}
\bibliography{references}

\end{document}